\documentclass{article}
\usepackage{iclr2022_workshop,times}
\usepackage{multirow}
\usepackage{enumitem}
\usepackage{graphics,subfig}

\usepackage{float}
\usepackage{diagbox}
\usepackage{array}
\usepackage{color}
\usepackage{graphicx} 
\usepackage{booktabs}
\usepackage{comment}
\usepackage{amsmath,amsfonts,bm}

\def\1{\bm{1}}

\DeclareMathAlphabet{\mathsfit}{\encodingdefault}{\sfdefault}{m}{sl}
\SetMathAlphabet{\mathsfit}{bold}{\encodingdefault}{\sfdefault}{bx}{n}

\iclrfinalcopy
\usepackage{hyperref}
\usepackage{url}

\title{Just Propagate: Unifying Matrix Factorization, Network Embedding, and LightGCN for Link Prediction}

\author{
Haoxin Liu\\
 \texttt{liuhaoxinthu@gmail.com}
}

\begin{document}

\maketitle

\begin{abstract}
Link prediction is a fundamental task in graph analysis. Despite the success of various graph-based machine learning models for link prediction, there lacks a general understanding of different models. In this paper, we propose a unified framework for link prediction that covers matrix factorization and representative network embedding and graph neural network methods. Our preliminary methodological and empirical analyses further reveal several key design factors based on our unified framework. 
We believe our results could deepen our understanding and inspire novel designs for link prediction methods. 
%The abstract paragraph should be indented 1/2~inch (3~picas) on both left and right-hand margins. Use 10~point type, with a vertical spacing of 11~points. The word \textsc{Abstract} must be centered, in small caps, and in point size 12. Two line spaces precede the abstract. The abstract must be limited to one paragraph.
\end{abstract}
\section{Introduction}
Link prediction, i.e.,  predicting the existence of links between nodes, is a fundamental task in graph data mining with important applications such as recommendation systems~\citep{pinsage}, protein interaction prediction~\citep{ppi},
knowledge discovery~\citep{kgsurvey}, etc. %anomaly detection~\citep{tao2019mvan}, etc. 
Many machine learning methods have been developed to tackle the link prediction problem~\citep{lp1,lp2}, ranging from early heuristic features~\citep{heu1,heu2}, statistical machine learning~\citep{line,deepwalk}, to the recent deep learning methods~\citep{graphsage,ngcf}. For example, matrix factorization (MF)~\citep{mf1} is a representative method for link prediction with the assumption that the graph structure has an underlying low-rank structure. Network embedding methods such as DeepWalk~\citep{deepwalk} and LINE~\citep{line} further consider the high-order proximity between nodes.
Recently, neural network-based graph learning methods such as NGCF~\citep{ngcf} and LightGCN~\citep{he2020lightgcn} have shown superior performance, which has been attributed to the explicit modeling of higher-order neighbors.

Despite the successes of the existing approaches, the aforementioned studies are relatively independent with different assumptions and design objectives, and there lacks a general understanding of the graph-based machine learning models for link prediction.
%a unified framework to help people better understand these methods.
Therefore, as the first contribution of this work, we propose a unified framework for link prediction methods, covering matrix factorization, network embedding methods such as DeepWalk~\citep{deepwalk} and LINE~\citep{line}, and graph neural network-based methods including LightGCN~\citep{he2020lightgcn}. Based on the unified framework, we further conduct in-depth methodological and empirical analyses. We make the following observations.
\begin{itemize}[leftmargin = 0.5cm]
    \item The overall learning process of the existing methods covered by our unified framework, including the forward inference and backward gradient descent for optimization, is equivalent to a representation propagation procedure. % based on the balance theory.
    \item Propagation kernels, which determine how the representation is propagated and whether higher-order neighbor information can be effectively utilized, greatly affect the performance of different methods for the link prediction task. 
    \item By regarding randomly drawn negative samples as negative edges, the existing approaches essentially follow the balance theory of signed graphs in aggregating neighborhoods.
\end{itemize}

We believe our proposed unified framework and preliminary analysis results could deepen our understanding and inspire novel designs for graph-based link prediction methods.
\section{Preliminaries}
\subsection{Notations}
Firstly, we introduce some notations used in the paper. Consider a graph $\mathcal{G}=\left(\mathcal{V},\mathcal{E} \right)$, where $\mathcal{V}$ denotes the node set and $\mathcal{E}$ denotes the link set. The adjacency matrix is denoted as $\mathbf{A}$. The goal of link prediction is to discover potential links from the observed links $\mathbf{A}$. 
We assume the graph is undirected and the nodes have no features. 
Notice that our problem setting naturally covers the recommendation problem, i.e., ${\mathcal{V}=\mathcal{U}\cup\mathcal{I}}$, and links only exist between $\mathcal{U}$ and $\mathcal{I}$.    

In the common setting of link prediction, there are no explicit negative samples~\cite{liu2023hapens}. For example, people seldom explicitly mark whom they do not want to be friends with or products they do not want to buy.  

Therefore, negative sampling is usually used in link prediction to sample from pairs of nodes that do not have links as pseudo negative samples during the optimization procedure. We denote the adjacency matrix for sampled negative links as $\mathbf{B} \in \left\{0,1\right\}^{|\mathcal{V}|\times|\mathcal{V}|}$, where $\mathbf{B}_{i,j}=1$ represents that node $i$ and $j$ has a negative link. We summarize other notations in Appendix~\ref{Notations}.

\subsection{Loss Function}

In this paper, we focus on a widely adopted loss function, the binary cross-entropy (BCE)  loss~\citep{neumf} and leave exploring other loss functions as future works. 
The basic BCE loss with a L2 regularization term is formulated as
\begin{equation}\label{eq:bceloss}
    \mathcal{L}_{\text{BCE}}=-\frac{1}{2}\sum \nolimits_{v \in \mathcal{V}}\left(\sum \nolimits_{i \in \mathcal{N}_A{(v)}} \log \sigma\left(\mathbf{X}_{v}^{\top} \mathbf{X}_{i}\right) +\lambda\sum \nolimits_{j \in \mathcal{N}_B(v)}\log \sigma\left(-\mathbf{X}_{v}^{\top} \mathbf{X}_{j}\right)\right)+\frac{\beta}{2}\|\mathbf{X}\|_2^{2},
\end{equation}
where $\mathcal{N}_A{(v)}$ and $\mathcal{N}_B{(v)}$ denote the positive and negative neighbors of node $v$, $\sigma(\cdot)$ is the Sigmoid activation function, $\lambda$ and $\beta$ are hyper-parameters to control the strength of the corresponding terms, and $\mathbf{X}$ denotes the representation of nodes.
Eq.~\eqref{eq:bceloss} can be written in an equivalent matrix form:
\begin{equation}
    \mathcal{L}_{\text{BCE}}=-\frac{1}{2}\mathbf{1}_{|\mathcal{V}|}\left(\left[\log \left(
    %\mathbf{1}+\exp \left(-\mathcal{Q}(\mathbf{X})\right)
    \sigma(\mathbf{X} \mathbf{X}^{\top})
    \right) \odot \mathbf{A}\right]+\lambda\left[\log \left(
    %\mathbf{1}+\exp \left(\mathcal{Q}(\mathbf{X})\right)
    -\sigma(\mathbf{X} \mathbf{X}^{\top})
    \right) \odot \mathbf{B}\right]\right)\mathbf{1}_{|\mathcal{V}|}^{\top}\\
    +\frac{\beta}{2}\|\mathbf{X}\|_2^{2},
\end{equation}
where %$\mathcal{Q}(\mathbf{X})=\mathbf{X} \mathbf{X}^{\top}$,
$\mathbf{1}_{|\mathcal{V}|}$ is a $1 \times|\mathcal{V}|$ row matrix with the values being ones and $\odot$ is the Hadamard product.
\subsection{A Brief Introduction to Considered Link Prediction Models}
In this section, we briefly introduce link prediction models considered in our framework. 
We consider four representative methods divided into three categories:
\begin{itemize}[leftmargin = 0.4cm]
    \item Matrix factorization, which directly optimizes the representation matrix $\mathbf{X}$ using the loss function.
    \item  DeepWalk and LINE: these network embedding methods also optimize the representation matrix $\mathbf{X}$, but with different loss functions that consider the high-order neighborhoods between nodes.
    \item  LightGCN: a linear GNN model that first propagates the representation as $\bar{\mathbf{X}}=\sum_{k=0}^{K}\frac{1}{K+1}{\mathbf{\tilde{A}}}^{k}\mathbf{X}$, $\tilde{\mathbf{A}}$ is a normalized adjacency matrix, e.g., $\tilde{\mathbf{A}}=\mathbf{D}_A^{-\frac{1}{2}}\mathbf{A}\mathbf{D}_A^{-\frac{1}{2}}$, $\mathbf{D}_A$ is the diagonal degree matrix, and then optimize $\mathbf{X}$ using the loss function.
\end{itemize}

Following the standard training paradigm in machine learning, we assume all these methods are optimized by the full-batch gradient descent and denote the node representations in the $m^{th}$ optimization step as $\mathbf{X}^{(m)}$, i.e., $\mathbf{X}^{(m+1)} = \mathbf{X}^{(m)} - \alpha \nabla_{\mathbf{X}^{(m)}}\mathcal{L}$.

\section{Our Proposed Framework}\label{sec:Framework}
\subsection{The Unified Framework}
Although the aforementioned methods appear to have unrelated design objectives and loss functions based on diverse motivations, we find that they can be unified into a general framework that only performs forward propagation, i.e., even the back propagation for gradient descends is equivalent to the forward propagation. Specifically, we have:

\begin{equation}\label{eq:unify}
\begin{aligned}
    \mathbf{X}^{(m+1)}&=\mathbf{H}^{(m+1)}\mathbf{X}^{(m)}\\
    &=\big(c_1\mathbf{I}+c_2\mathcal{P}_{a_1,b_1}(\tilde{\mathbf{A}})\left[\mathcal{K}_{+}^{(m+1)}(\mathbf{A})-\lambda\mathcal{K}_{-}^{(m+1)}(\mathbf{B})\right]\mathcal{P}_{a_1,b_1}(\tilde{\mathbf{A}})\big)\mathbf{X}^{(m)},
\end{aligned}
\end{equation}
where $c_1, c_2$ are constants that depend on the learning rate $\alpha$ and regularization term $\beta$, $a_1,b_1 \in \mathbb{N}$ are model-specific constants, and $\mathbf{I}$ is the $|V|\times|V|$ identity matrix.
\begin{equation}\label{eq:unify}
\begin{aligned}
    \mathbf{X}^{(m+1)}&=\mathbf{H}^{(m+1)}\mathbf{X}^{(m)}\\
    &=\big(c_1\mathbf{I}+c_2\mathcal{P}_{a_1,b_1}(\tilde{\mathbf{A}})\left[\mathcal{K}_{+}^{(m+1)}(\mathbf{A})-\lambda\mathcal{K}_{-}^{(m+1)}(\mathbf{B})\right]\mathcal{P}_{a_1,b_1}(\tilde{\mathbf{A}})\big)\mathbf{X}^{(m)}\\
    &=\big(c_1\mathbf{I}+c_2\left[\mathcal{K}_{+}^{(m+1)}(\mathbf{A})-\lambda\mathcal{K}_{-}^{(m+1)}(\mathbf{B})\right]\big)\mathbf{X}^{(m)}\\
    &=\big(c_1\mathbf{I}+c_2\left[\mathbf{S}_A^{(m+1)}\odot\tilde{\mathbf{A}}-\lambda\mathbf{S}_B^{(m+1)}\odot\tilde{\mathbf{B}}\right]\big)\mathbf{X}^{(m)}
\end{aligned}
\end{equation}
The detailed proof and the correspondence of different methods to our framework are provided in Appendix~\ref{sec:proof} and Appendix~\ref{Details}, respectively. Though Eq.~\eqref{eq:unify} appears to be complicated, we elaborate the terms one by one. 

Firstly, $\mathbf{H}^{(m+1)}\in\mathbb{R}^{|\mathcal{V}|\times|\mathcal{V}|}$ is the propagation kernel that determines how the representation is propagated. It is composed of three major terms: a high-order proximity matrix $\mathcal{P}_{a_1,b_1}(\tilde{\mathbf{A}})$, a positive link kernel $\mathcal{K}_{+}^{(m+1)}(\mathbf{A})$, and a negative link kernel $\mathcal{K}_{-}^{(m+1)}(\mathbf{B})$.
The high-order matrix is defined as:
\begin{equation}
    \mathcal{P}_{a,b}(\tilde{\mathbf{A}})=\sum \nolimits_{r=a}^{b}\frac{1}{b-a+1}\tilde{\mathbf{A}}^{r},
\end{equation} 
i.e., it is a weighted sum of the $a^{th}$ order to the $b^{th}$ order normalized adjacency matrix $\tilde{\mathbf{A}}$, $a \leq b$.
The positive and negative link kernels are defined as follows:
\begin{equation}\label{score}
\begin{aligned}
\mathcal{K}_{+}^{(m+1)}(\mathbf{A})&=\mathbf{S}_A^{(m+1)}\odot(c_3\mathcal{P}_{a_2,b_2}(\tilde{\mathbf{A}})+(1-c_3)\mathbf{A})\\
\mathcal{K}_{-}^{(m+1)}(\mathbf{B})&=\mathbf{S}_B^{(m+1)}\odot(c_3\tilde{\mathbf{B}}+(1-c_3)\mathbf{B}),
\end{aligned}
\end{equation}
where $c_3 \in\{0,1\}$, $\tilde{\mathbf{B}}$ is the normalized adjacency matrix of negative links, e.g., $\tilde{\mathbf{B}}=\mathbf{D}_B^{-\frac{1}{2}}\mathbf{B}\mathbf{D}_B^{-\frac{1}{2}}$, and $\mathbf{S}_A^{(m+1)}, \mathbf{S}_B^{(m+1)}$ are score matrices that measure the difference between the similarity calculated by the representation vectors and the target link as follows
%. Since these methods are all based on inner product, their corresponding iterative scored matrices can be specialized as the following form:
\begin{equation}
    \begin{aligned}\label{eq:LGC_PRO}
      \mathbf{S}_A^{(m+1)}&=1-\sigma\left(
      %\mathcal{Q}(\mathbf{\bar{X}^{(m)}})
      \bar{\mathbf{X}}^{(m)} \bar{\mathbf{X}}^{(m)\top}
      \right), \mathbf{S}_B^{(m+1)}=\sigma\left(
      %\mathcal{Q}(\mathbf{\bar{X}^{(m)}})
      \bar{\mathbf{X}}^{(m)} \bar{\mathbf{X}}^{(m)\top}
      \right),
    \end{aligned}
\end{equation}
where $\mathbf{\bar{X}}^{(m)}=\mathcal{P}_{a_1,b_1}(\tilde{\mathbf{A}})\mathbf{X}^{(m)}=\sum_{r=a_1}^{b_1}\frac{1}{b_1-a_1+1}\tilde{\mathbf{A}}^{r}\mathbf{X}^{(m)}$ is the propagated node representation.
From Eq.~\eqref{eq:LGC_PRO}, score matrices equal $0$ when the representation perfectly reconstructs the training data.

In summary, all these four representative methods can be unified in Eq.~\eqref{eq:unify}, which indicates that their underlying mechanism shares strong connections. In the next section, we conduct in-depth analyses based on this framework.

\subsection{Revisiting the Existing Link Prediction Methods}
In this section, we revisit the existing methods based on our proposed framework.

Firstly, it is easy to see that all the four covered methods have analytical solutions only using the forward propagation, i.e., doing back propagation for gradient descends ends up being equivalent to changing the forward propagation kernel. Therefore, our framework sheds lights on novel design principles for link prediction that do no involve computationally expensive back propagations. 

In view of signed graphs~\cite{liu2021signed,liu2022lightsgcn} that sampled negative links correspond to real negative connections, the propagation step in Eq.~\eqref{eq:unify} satisfies the balance theory, i.e., a sociological assumption that ``the enemy of my enemy is my friend''. Though this assumption is never explicitly stated for any of the covered methods, it is always behind the curtain and  naturally surfaces from our unified framework, which should draw further research attention.

Different methods have diverse strategies in using the high-order information from neighborhoods. To be more specific, from Eq.~\eqref{eq:unify}, it is easy to see that both $a_1,b_1$ in the high-order proximity matrix $\mathcal{P}_{a_1,b_1}(\tilde{\mathbf{A}})$ and $a_2,b_2$ in the positive link kernel $\mathcal{K}_{+}^{(m+1)}(\mathbf{A})$ affects the high-order proximity (please see Appendix~\ref{Details} for the exact values of different methods). For MF and LINE, they can aggregate one extra order proximity during each optimization step. Therefore, MF and LINE can learn from high-order neighborhoods by accumulating steps during optimization. Such a result is in sharp contrast with the current belief that MF and LINE can only preserve the first and second-order proximity between nodes. In addition,
DeepWalk and LightGCN can aggregate multiple orders of neighbors during each optimization step through two separate mechanisms: DeepWalk constructs a high-order proximity in the objective function using random walks, while LightGCN contains an internal propagation step. In short, how different methods utilize high-order proximities is vital to the model design.

\begin{figure}[t]
\centering
\includegraphics[width=14cm]{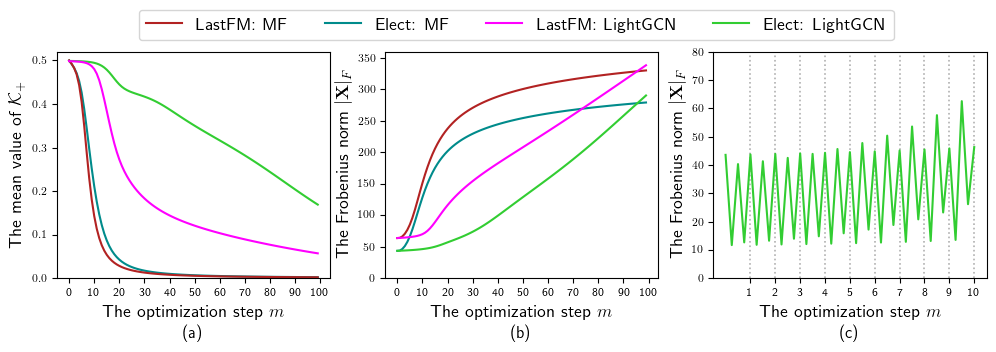}
%\caption{fig1}
\centering
\vspace*{-0.5cm}
\caption{The analysis results of MF and LightGCN. (a) The mean value of $\mathcal{K}_+$ as the optimization step increases; (b) The Frobenius norm of $\mathbf{X}$ as the optimization step increases; (c) The Frobenius norm  $\left\| \mathbf{X} \right\|_F$ for LightGCN as the optimization substeps increase. }
\vspace*{-0.5cm}
%\caption{The variation of $\overline{\mathcal{K}_{+}^{(m)}(\mathbf{A})}$, i.e., the mean score of $\mathcal{K}_{+}^{(m)}(\mathbf{A})$ and $\|\mathbf{X}^{(m)}\|_2$, i.e, the frobenius norm of $\mathbf{X}^{(m)}$. (a) }
\label{fig:exp}
\end{figure}

\iffalse
\begin{figure}[t]
\centering
\subfigure[]{
\centering
\includegraphics[width=4.5cm]{picture/mean_n3.png}
%\caption{fig1}
}\hspace{-0.1cm}
\subfigure[]{
\centering
\includegraphics[width=4.5cm]{picture/norm_3.png}\hspace{-0.1cm}
%\caption{fig2}
}
\centering
\centering
\subfigure[]{
\centering
\includegraphics[width=4.5cm]{picture/A_norm_step_elect.png}\hspace{-0.1cm}
%\caption{fig1}
}
\centering
\caption{The analysis results of MF and LightGCN. (a) The mean value of $\mathcal{K}_+$ as the optimization step increases; (b) The Frobenius norm of $\mathbf{X}$ as the optimization step increases; (c) The Frobenius norm  $\left\| \mathbf{X} \right\|_F$ for LightGCN as the optimization substeps increase. }
\label{fig:exp}
\end{figure}
\fi

To empirically analyze different methods based on our framework, we conduct experiments on two recommendation datasets: Elect and LastFM.

We mainly compare the results of MF and LightGCN with some analyses.
More details about the experimental settings can be found in Appendix~\ref{settings}. 

The results are shown in Appendix~\ref{results}. Not surprisingly, LightGCN achieves better results than MF, which is consistent with the literature. Based on the results, we try to investigate a more fundamental question: "What factors drive the success of link prediction models?" Using our framework, we preliminary examine two targets: the positive link kernel $\mathcal{K}_+$ and the representation matrix $\mathbf{X}$.

For the positive link kernel, recall that it measures how well the node representation can reconstruct the positive links. Therefore, its average value is a good indicator of the model learning process. To analyze its influence, we plot the mean value of $\mathcal{K}_+$ as the optimization progresses in Figure~\ref{fig:exp}(a). 
The figure clearly shows that the mean values for MF drop much faster than LightGCN, indicating that MF quickly fits the training positive data. However, the final results of LightGCN are much better (both methods have already adopted early stopping to prevent overfitting). Therefore, the empirical results show that we need to properly control the fitting curve to realize effective link prediction models. 

Next, we analyze the representation matrix $\mathbf{X}$. Specifically, since inner product is used to measure the similarity between node representations, both the length and direction of representation vectors affect the similarity. Here, we focus on the average length of representation vectors and plot the Frobenius norm of $\mathbf{X}$ vs. the optimization steps in Figure~\ref{fig:exp}(b). The results indicate that the representation of LightGCN has smoother lengths than MF. To further analyze the underlying reason, we decompose one optimization step of LightGCN, i.e., one step of Eq.~\eqref{eq:unify}, into four substeps   (please check  Appendix~\ref{Details} for details):
(1) Propagate $K$ layers using $\tilde{\mathbf{A}}$; (2) Propagate one layer using $\mathcal{K}_{+}^{(m+1)}(\mathbf{A})$; (3) Propagate another $K$ layers using $\tilde{\mathbf{A}}$; (4) Update the representations. 

We further plot the Frobenius norm of LightGCN with respect to these four substeps in Figure~\ref{fig:exp}(c). The results show that substeps (1) and (3) can reduce the length while steps (2) and (4) increase the length. The reason is that the propagations in (1) and (3) use the normalized adjacency matrix $\tilde{\mathbf{A}}$, which can control the length of the representation.
The results show that properly choosing the normalization of adjacency matrix is crucial for the success of link prediction methods. 

\section{Conclusion}
In this paper, we introduce a unified framework for four representative link prediction methods including MF, DeepWalk, LINE, and LightGCN. We reveal several key design principles of these methods based on our proposed framework. We believe our proposed framework and in-depth analyses can open up new research opportunities for developing link prediction methods.  %containing their whole process of representation learning. This framework demonstrates that they are actually performing propagation with corresponding kernels. It also allows us to rethink the success of graph based-CF thanks to it's good interpretability. We empirically show that the combination of two weighted adjancency matrix can ultilize the high-order neighbor information effectively and  have a significant impact on the performance, rather than the common belief of constrcuting graphs and propagation explicitly. We believe our framework. Our proposed unified framework and conclusions,  can also opens up new opportunities for designing new CF methods.
\bibliography{iclr2022_workshop}
\bibliographystyle{iclr2022_workshop}
\appendix
\section{Main Notations}\label{Notations}
\begin{table}[H]
\begin{center}
	\caption{Notations used in the paper}
	\label{tab:freq}
	\resizebox{\linewidth}{!}{
	\begin{tabular}{ll}
		\toprule
		Notations     &  Descriptions     \\
		\midrule
		%$\mathbf{R}\in \mathbb{R}^{|\mathcal{U}|\times|\mathcal{I}|}$ &  The original implicit feedback matrix\\
		$\mathbf{A}\in \left\{0,1 \right\}^{|\mathcal{V}|\times|\mathcal{V}|}$ &  The adjacency matrix of positive links  \\
		$\mathbf{B}\in \left\{0,1 \right\}^{|\mathcal{V}|\times|\mathcal{V}|}$ &  The adjacency matrix of negative links by negative sampling  \\
		$\mathbf{D}_{A}/\mathbf{D}_{B}$ & The diagonal degree matrix of $\mathbf{A}/\mathbf{B}$
		\\
		%$\mathcal{U,I,V}$ &  A set of uses, items and all nodes\\
		$\mathcal{N}_A{(u)}$ & The positive neighborhood of node $u$\\
		$\mathcal{N}_B{(u)}$ & The sampled negative neighborhood of node $u$\\
		$\mathbf{X}^{(m)}$  &   The node representation matrix at the $m^{th}$ epoch \\
		%$\mathbf{Z}$ &  The final representation matrix\\
		%$K$           &   The number of propagation layers in GNN-based modules\\
		%$M$           &   The number of training epochs\\
		$\alpha$           &   The learning rate\\
        $\beta$           &   The strength for the $L_2$ regularization \\
        $\mathbf{H}^{(m)}$  & The propagation kernel at the $m^{th}$ epoch \\
        $\tilde{\mathbf{A}}(\tilde{\mathbf{B}})$	& The normalized adjacency matrix of positive and negative links \\
        $\mathbf{S}_{A}^{(m)}/\mathbf{S}_{B}^{(m)}$	&The scoring matrix for positive and negative links at the $m^{th}$ epoch \\
        
        $\sigma(\cdot)$ &The Sigmoid activation function\\
         $\odot$ & The Hadamard product\\
    \bottomrule
\end{tabular}}
\end{center}
\end{table}
\section{Proof}\label{sec:proof}
In this section, we first give the modified loss function of the corresponding methods in the matrix format. Then, we take LightGCN as an example to show the derivation of our unified framework, while other methods can be proved in similar ways. 
\subsection{Loss}
For matrix factorization, a commonly used loss function is the original BCE loss, i.e., $\mathcal{L}_{\text{MF}}=\mathcal{L}_{\text{BCE}}$.
For LINE,
%Considering edge direction and weight, LINE with the second order proximity (aka LINE (2nd)) aims to learn two representation matrices: the vertices representation $\mathbf{X}\in\mathcal{R}^{|\mathcal{V}|\times d}$, and the context representation $\mathbf{Y}\in\mathcal{R}^{|\mathcal{V}|\times d}$. 
its loss function can be written as follows
\begin{equation}
     \mathcal{L}_{\text{LINE}}=-\frac{1}{2}\mathbf{1}_{|\mathcal{V}|}\left(\left[\log \left(
    %\mathbf{1}+\exp \left(-\mathcal{Q}(\mathbf{X})\right)
    \sigma(\mathbf{X} \mathbf{X}^{\top})
    \right) \odot \tilde{\mathbf{A}}\right]+\lambda\left[\log \left(
    %\mathbf{1}+\exp \left(\mathcal{Q}(\mathbf{X})\right)
    -\sigma(\mathbf{X} \mathbf{X}^{\top})
    \right) \odot \mathbf{B}\right]\right)\mathbf{1}_{|\mathcal{V}|}^{\top}\\
    +\frac{\beta}{2}\|\mathbf{X}\|_2^{2},
\end{equation}
where $\tilde{\mathbf{A}}=\mathbf{D}_A^{-1}\mathbf{A}$ and $\mathbf{B}$ is the  expectation of negative sampled links as follows
\begin{equation}
    \mathbf{B} \propto\mathbf{D}_A^{\frac{3}{4}}.%\mathbf{1}_{|\mathcal{V}|\times|\mathcal{V}|}.
\end{equation}
%where $(\mathbf{D}_A^{\frac{3}{4}})^{'}\propto\mathbf{D}_A^{\frac{3}{4}}$ in the original paper.
%Since only undirected and unweighted graphs are considered in this paper, we assum $\mathbf{X}=\mathbf{Y}$ and adopt negative sampling methods. Then we rewrite $\mathcal{L}_{\text{LINE}}$ as
%\begin{equation}
%\begin{aligned}
%    \mathcal{L}_{\text{LINE}}=-\frac{1}{2}\mathbf{1}_{|\mathcal{V}|}\left(\left[\log \left(
    %\mathbf{1}+\exp \left(-\mathcal{Q}(\mathbf{X})\right)
%    \sigma(\mathbf{X} \mathbf{X}^{\top})
%    \right) \odot \tilde{\mathbf{A}}\right]+\lambda\left[\log \left(
    %\mathbf{1}+\exp \left(\mathcal{Q}(\mathbf{X})\right)
%    -\sigma(\mathbf{X} \mathbf{X}^{\top})
%    \right) \odot \tilde{\mathbf{B}}\right]\right)\mathbf{1}_{|\mathcal{V}|}^{\top}+\frac{\beta}{2}\left(\|\mathbf{X}\|_2^{2}\right)
%\end{aligned}
%\end{equation}
%where $\tilde{\mathbf{B}}=(\mathbf{D}_A^{\frac{3}{4}})^{'}\mathbf{B}$. 
For DeepWalk, 
based on the analysis by \citet{qiu2018network}, when the walk length $L\rightarrow +\infty$, the loss function of DeepWalk in the matrix form with the window size $w\in\mathbb{N}^{+}$ can be written as 
\begin{equation}
    \mathcal{L}_{\text{DW}}=-\frac{1}{2}\mathbf{1}_{|\mathcal{V}|}\left(\left[\log \left(
    %\mathbf{1}+\exp \left(-\mathcal{Q}(\mathbf{X})\right)
    \sigma(\mathbf{X} \mathbf{X}^{\top})
    \right) \odot \mathcal{P}_{1,w}(\tilde{\mathbf{A}})\right]+\lambda\left[\log \left(
    %\mathbf{1}+\exp \left(\mathcal{Q}(\mathbf{X})\right)
    -\sigma(\mathbf{X} \mathbf{X}^{\top})
    \right) \odot \tilde{\mathbf{B}}\right]\right)\mathbf{1}_{|\mathcal{V}|}^{\top}
    +\frac{\beta}{2}\|\mathbf{X}\|_2^{2},
\end{equation}
where $\tilde{\mathbf{A}}=\mathbf{D}_A^{-1}\mathbf{A}$ and $\tilde{\mathbf{B}}=\mathbf{D}_B^{-1}\mathbf{B}$. 

As for LightGCN, the loss function can be written in a matrix form as
\begin{equation}\label{eq:LGC}
    \mathcal{L}_{\text{LGC}}=-\frac{1}{2}\mathbf{1}_{|\mathcal{V}|}\left(\left[\log \left(
    %\mathbf{1}+\exp \left(-\mathcal{Q}(\mathbf{X})\right)
    \sigma(\mathbf{\bar{X}} \mathbf{\bar{X}}^{\top})
    \right) \odot \mathbf{A}\right]+\lambda\left[\log \left(
    %\mathbf{1}+\exp \left(\mathcal{Q}(\mathbf{X})\right)
    -\sigma(\mathbf{\bar{X}} \mathbf{\bar{X}}^{\top})
    \right) \odot \mathbf{B}\right]\right)\mathbf{1}_{|\mathcal{V}|}^{\top}\\
    +\frac{\beta}{2}\|\mathbf{X}\|_2^{2}
\end{equation}
where $\bar{\mathbf{X}}=\sum_{k=0}^{K}\frac{1}{K+1}{\tilde{\mathbf{A}}}^{k}\mathbf{X}$, $\tilde{\mathbf{A}}=\mathbf{D}_A^{-\frac{1}{2}}\mathbf{A}\mathbf{D}_A^{-\frac{1}{2}}$.
\subsection{LightGCN Derivation}
In this subsection, we take LightGCN as an example to show how to derive our unified framework. The derivative of Eq.~\eqref{eq:LGC} is calculated as
\begin{equation}
    \begin{aligned}
         \frac{\partial\mathcal{L}_{\text{LGC}}}{\partial \mathbf{X}^{(m)}}&=\left(\beta\mathbf{I}- \sum_{k=0}^{K}\frac{{\tilde{\mathbf{A}}}^{k}}{K+1}
         \Big[\big(1-\sigma(
      \bar{\mathbf{X}}^{(m)} \bar{\mathbf{X}}^{(m)\top}
      )\big)\odot\mathbf{A} -\lambda\sigma(
      \bar{\mathbf{X}}^{(m)} \bar{\mathbf{X}}^{(m)\top}
      )\odot\mathbf{B}\Big]\sum_{k^\prime=0}^{K}\frac{{\tilde{\mathbf{A}}}^{k^\prime}}{K+1}\right)\mathbf{X}^{(m)}\\
      &=\Big(\beta\mathbf{I}-\mathcal{P}_{1,K}(\tilde{\mathbf{A}})\left[\mathbf{S}_{A}^{(m+1)}\odot\mathbf{A}-\lambda\mathbf{S}_{B}^{(m+1)}\odot\mathbf{B}\right]\mathcal{P}_{1,K}(\tilde{\mathbf{A}})\Big)\mathbf{X}^{(m)}.
      %&=\Big(\beta\mathbf{I}-\mathcal{P}_{1,K}(\tilde{\mathbf{A}})\left[\mathcal{K}_{+}^{(m+1)}(\mathbf{A})-\lambda\mathcal{K}_{-}^{(m+1)}(\mathbf{B})\right]\mathcal{P}_{a_1,b_1}(\tilde{\mathbf{A}})\Big)\mathbf{X}^{(m)}
    \end{aligned}
\end{equation}
Then, we have
\begin{equation}
    \begin{aligned}
         \mathbf{X}^{(m+1)}&=\mathbf{X}^{(m)}-\alpha\frac{\partial\mathcal{L}_{\text{LGC}}}{\partial \mathbf{X}^{(m)}}\\
         &=\Big((1-\alpha\beta)\mathbf{I}+\alpha\mathcal{P}_{1,K}(\tilde{\mathbf{A}})\left[\mathbf{S}_{A}^{(m+1)}\odot\mathbf{A}-\lambda\mathbf{S}_{B}^{(m+1)}\odot\mathbf{B}\right]\mathcal{P}_{1,K}(\tilde{\mathbf{A}})\Big)\mathbf{X}^{(m)},
    \end{aligned}
\end{equation}
which is a special case of Eq.~\eqref{eq:unify} by 
setting $c_1=1-\alpha\beta$, $c_2=\alpha$, $a_1=a_2=b_2=0$, $b_1=K$ and $\tilde{\mathbf{A}}=\mathbf{D}_A^{-\frac{1}{2}}\mathbf{A}\mathbf{D}_A^{-\frac{1}{2}}$.$\tilde{\mathbf{A}}=\mathbf{D}_A^{-1}\mathbf{A}$.$\tilde{\mathbf{B}}=\mathbf{D}_B^{-1}\mathbf{B}$.
\begin{equation}
    \begin{aligned}
         \mathbf{H}^{(m+1)}_{\text{LGC}}=(1-\alpha\beta)\mathbf{I}+\alpha\mathcal{P}_{1,K}(\tilde{\mathbf{A}})\left[\mathbf{S}_{A}^{(m+1)}\odot\mathbf{A}-\lambda\mathbf{S}_{B}^{(m+1)}\odot\mathbf{B}\right]\mathcal{P}_{1,K}(\tilde{\mathbf{A}})
    \end{aligned}
\end{equation}
\begin{equation}
    \begin{aligned}
         \mathbf{H}^{(m+1)}_{\text{MF}}=(1-\alpha\beta)\mathbf{I}+\left[\mathbf{S}_{A}^{(m+1)}\odot\mathbf{A}-\lambda\mathbf{S}_{B}^{(m+1)}\odot\mathbf{B}\right]
    \end{aligned}
\end{equation}
%the unified framework in Eq.~\eqref{eq:unify} can be specialized as
%\begin{equation}
%    \begin{aligned}
 %   \mathbf{X}^{(m+1)}&=\mathbf{H}^{(m+1)}\mathbf{X}^{(m)}\\
%    &=\big((1-\alpha\beta)\mathbf{I}+\alpha\mathcal{P}_{1,K}(\tilde{\mathbf{A}})\left[\mathcal{K}_{+}^{(m+1)}(\mathbf{A})-\lambda\mathcal{K}_{-}^{(m+1)}(\mathbf{B})\right]\mathcal{P}_{1,K}(\tilde{\mathbf{A}})\big)\mathbf{X}^{(m)}\\
%    &=\Big(\beta\mathbf{I}-\mathcal{P}_{1,K}(\tilde{\mathbf{A}})\left[\mathbf{S}_{A}^{(m+1)}\odot\mathbf{A}-\lambda\mathbf{S}_{B}^{(m+1)}\odot\mathbf{B}\right]\mathcal{P}_{1,K}(\tilde{\mathbf{A}})\Big)\mathbf{X}^{(m)}
%\end{aligned}
%\end{equation}
\section{Detailed Correspondence of Link Prediction Models }\label{Details}

We summarize the detailed correspondence of different models in our unified framework in Table~\ref{table:correspondence}.

\begin{table}[h]
\renewcommand\arraystretch{1.8}
\begin{center}
\caption{The detailed correspondence of different models in our unified framework. }\label{table:correspondence}
\begin{tabular}{|c|c|c|c|c|c|c|c|c|c|}
\hline
Model &$c_1$ & $c_2$ & $c_3$ &$a_1$ &$b_1$&$a_2$ &$b_2$& $\tilde{\mathbf{A}}$ & $\tilde{\mathbf{B}}$ \\ \hline
      MF&$1-\alpha\beta$   &$\alpha$   &0   &0 &0 &0&0& --- & ---      \\ \hline
      LINE&$1-\alpha\beta$   &$\alpha$   &1 &0 &0 &1&1    &$\mathbf{D}_A^{-1}\mathbf{A}$ &$\propto \mathbf{D}_A^{\frac{3}{4}}\mathbf{B}$    \\ \hline
      DeepWalk&$1-\alpha\beta$   &$\alpha$   &1   &0   &0 &1&$w$&$\mathbf{D}_A^{-1}\mathbf{A}$ &$\mathbf{D}_B^{-1}\mathbf{A}$        \\ \hline
      LightGCN&$1-\alpha\beta$   &$\alpha$  &0   &0 &$K$&0&0 &$\mathbf{D}_A^{-\frac{1}{2}}\mathbf{A}\mathbf{D}_A^{-\frac{1}{2}}$& ---   \\
      \hline
\end{tabular}
\end{center}
\end{table}
\section{Experimental Settings}\label{settings}
\subsection{Datasets}
We adopt the following two public datasets in our experiments:
\begin{itemize}[leftmargin=0.5cm]
\item Elect is a dataset from the Amazon-review collection about electronics, which is widely used for item recommendation tasks. We use the dataset processed by~\citet{yu2020sampler}.
\item LastFM is collected from the music community site. It records the download relationships between users and artists. We use the dataset processed by~\citet{he2020lightgcn}.
\end{itemize}
The statistics of the datasets are summarized in Table~\ref{tab:statistics}

\begin{table}[ht]
\begin{center}
	\caption{The statistics of datasets}
	\label{tab:statistics}
	
	\begin{tabular}{cccc}\toprule
		Dataset &\# Nodes  &\# Links & Density\\ \midrule
		\texttt Elect      & 2,957 &35,931 &1.645\% \\
		\texttt LastFM      &6,381 &52,668 &0.620\% \\
		\bottomrule
	\end{tabular}
\end{center}
\end{table}

\subsection{Evaluation Metrics}
Following previous works~\citep{ngcf,he2020lightgcn}, we treat all non-interacted products as candidate links and output the link prediction scores and perform the top-K recommendation task, where $K$ is set as 20. As for the evaluation metrics, we choose Precision@K, Recall@K, and NDCG@K. For all the metrics, higher values indicate better performance. We repeat all experiments five times with different random seeds and report the average results. 

\subsection{Hyper-parameters}
We tune the learning rate $\alpha$ in $\{1e-5,1e-4,1e-3,1e-2,1e-1\}$ and the propagation layers for LightGCN in $\{1,3,5\}$. For each observed user-item interaction, we randomly sample
one negative pair from non-interacted items before training. Besides, we set $\lambda=1$ for all models. 
For all methods, we adopt an early stopping strategy with a patience of 10, i.e., stop the optimization if the validation accuracy does not improve for 10 steps.  

\section{Experimental Results}\label{results}
\begin{table}[H]
    		\centering
    		\caption{The result of link prediction for different methods.} \label{tab:revisting}
    		\begin{tabular}{|c |c |c |c |}
    			\hline {Dataset} & {Metric} &{MF}%&{LinearProp} 
    			&{LightGCN} \\
    			\hline
    			
    			\multirow{3}{*}{Electronics} 
    			& {Precision@20} & {0.0079} %&{0.0095} 
    			& {0.0157}  \\
    			\cline{2-4}
    			& {Recall@20} &
    		{0.0547} &
    		%{0.0696}&
    	 {0.1125} \\
    			\cline{2-4}
    			& {NDCG@20} & {0.0321} &
    			%{0.0393}&
    		 {0.0652} \\
    			
    			\hline
    			
    			\multirow{3}{*}{LastFM} 
    			& {Precision@20} & {0.0302} &
    			%{0.0363}&
    			 {0.0557}  \\
    			\cline{2-4}
    			& {Recall@20} & {0.1112} &
    			%{0.1316}&
    			 {0.1981} \\
    			\cline{2-4}
    			& {NDCG@20} & {0.0767} &
    			%{0.0852}&
    			 {0.1468} \\
    			\hline
    		\end{tabular}
    		\label{tab:abl_module}
    	\end{table}
\end{document}